\documentclass[10pt]{article}

\usepackage[letterpaper,margin=1in]{geometry}
\usepackage{newtxtext,newtxmath}
\usepackage{amsmath}

\usepackage{setspace}
\usepackage{cite}
\usepackage{xurl}

\usepackage{graphicx}
\usepackage{booktabs}
\usepackage{caption}
\usepackage{capt-of}
\usepackage{subcaption}
\usepackage{placeins}
\usepackage{longtable}
\usepackage{paracol}

\DeclareCaptionFont{tenpt}{\fontsize{10}{12}\selectfont}


\usepackage{titlesec}
\titleformat{\section}
  {\bfseries\fontsize{12}{14}\selectfont}
  {\thesection.}{0.5em}{}
\titlespacing*{\section}{0pt}{\baselineskip}{0pt}

\titleformat{\subsection}
  {\bfseries\fontsize{10}{12}\selectfont}
  {\thesubsection.}{0.5em}{}
\titlespacing*{\subsection}{0pt}{\baselineskip}{0pt}

\usepackage{fancyhdr}
\fancypagestyle{firstpage}{
  \fancyhf{}
  
  \fancyhead[L]{\fontsize{10}{12}\selectfont\itshape
    Proceedings of the IISE Annual Conference \& Expo 2026\\
    Y.\ Xiang, D.\ Yu, R.\ Thiesing, eds.}
}

\newcommand{\papertitle}{Data-Driven Resource Aggregation for Scalable Digital Twin and Simulation Model Generation}
\newcommand{\abstractid}{16838} 

\begin{document}
\thispagestyle{firstpage}

\begin{center}
{\bfseries\fontsize{16}{18}\selectfont \papertitle\par}
{\bfseries\fontsize{12}{14}\selectfont Abstract ID: \abstractid\par}
Yucheng Wu, Alp Akcay \\
Department of Mechanical and Industrial Engineering, Northeastern University, Boston, USA
\end{center}

\vspace{-1.0\baselineskip}

\begin{center}
{\bfseries\fontsize{12}{14}\selectfont Abstract\par}
\end{center}

{\fontsize{10}{12}\selectfont
As industrial digitization accelerates, the use of digital twins has become widespread. However, modern manufacturing systems are highly dynamic and frequently reconfigured, making manual model development and updates increasingly difficult. Recent research on automated simulation model generation shows that historical data such as event logs can yield simulation models automatically. However, modeling every resource at full granularity leads to unnecessary complexity and computational cost. We propose a method for automatically selecting and aggregating resources to simplify simulation modeling. Our method uses ideas from inverse reinforcement learning to learn behavioral value representations of resources from production event logs. This process identifies and aggregates resources with similar functional contributions. Experiments on manufacturing systems, using throughput and system time as performance measures, demonstrate that value-based resource aggregation performs well when the goal is to substantially reduce complexity while preserving performance accuracy.
\par}

\vspace{-1.0\baselineskip}

\begin{center}
{\bfseries\fontsize{12}{14}\selectfont Keywords\par}
\end{center}

{\fontsize{10}{12}\selectfont
Simulation; manufacturing systems;  resource aggregation; inverse reinforcement learning; machine learning
\par}

\section{Introduction}
Digital twin technology has emerged as a high-fidelity, real-time paradigm for simulating complex systems. By maintaining an up-to-date simulation model of the physical system, digital twins enable trend prediction, anomaly detection and predictive decision support in domains ranging from manufacturing to transportation\cite{kritzinger2018digital, tao2018digital}. However, building and maintaining simulation models of large-scale systems present significant challenges in terms of scalability and fidelity\cite{turner2016discrete, 10407831}. As system size and complexity increase, detailed simulation of every resource in the system can lead to excessive computational costs and model bloat. Conversely, overly coarse models may compromise fidelity\cite{jain2017manufacturing, fowler2004grand}. 

In this work, we address the question of how to automatically construct the simplest simulation model that preserves an accurate representation of system-level performance. Our focus is a flexible flow shop (FFS) manufacturing system~\cite{lee2019review, son2001automatic} controlled by a dispatching policy that assigns production jobs to resources. We utilize event logs from FFS to (1) learn the system's control policy, (2) identify appropriate resource aggregation schemes and (3) validate whether the simplified model can reproduce the key performance indicators (KPIs) of the original system. 


The remainder of the paper is organized as follows. Section~\ref{sec:lit} provides an overview of related work on automated simulation model generation. Section~\ref{sec: problem} describes an FFS system and formulates its evolution over time as a Markov decision process (MDP). Section~\ref{sec: method} introduces our proposed method, which leverages this MDP formulation and inverse reinforcement learning to learn a simulation model directly from event logs.  Section~\ref{sec: results} presents the numerical experiments, and Section~\ref{sec: conc} concludes the paper.

\section{Related Work}
\label{sec:lit}

Over the past two decades, various approaches have been proposed to achieve automated or self-updating simulation models, particularly in manufacturing systems. 
Mueller et al.\cite{mueller2007automatic} used Petri net data structures to automatically generate large-scale simulation models. Sommer et al.\cite{sommer2020improvement} achieved automated simulation model generation by performing object recognition and using predefined planning frameworks for the factory. Tan et al.\cite{tan2023automatic} applied process mining and iteratively generated improved simulation models using new data. Reinhardt et al.\cite{reinhardt2019survey} proposed a new classification of ASMG methods based on data input attributes. More recent studies have focused on improving the autonomy and intelligence of simulation models. Vernickel et al.\cite{vernickel2020machine} proposed using machine learning to identify specific parameters of processes in production systems. Pawletta and Bartelt\cite{pawletta2023integrating} explored how to integrate discrete events into a reinforcement learning environment, where the environment is formulated as a Markov decision process. More recently, Krenczyk\cite{krenczyk2024deep} used deep reinforcement learning methods to explore process allocation strategies and build production system models. Cimino et al.\cite{cimino2025automatic} systematically reviewed the development and application of so-called Automated Simulation Model Generation (ASMG) in manufacturing systems and emphasized that, although the concept of ASMG is not new, robust and widely applicable implementations are still at an early stage of development.

As the scale of systems addressed by ASMG increases, the idea of simplifying or aggregating parts of a model becomes increasingly attractive. Choudhary et al.\cite{choudhary2009data} pointed out that clustering is an important data mining function and resource clustering using real production data is an area that requires further research for simulation models. Some studies address this problem by focusing on specific parts of the system and using higher-level approximations. Deenen et al.\cite{deenen2024data} proposed a data-driven aggregation model for wafer fabrication. They combined functional areas such as lithography and etching into a single server and calibrated the service time distribution based on the actual throughput of that area. This method allows for accurate estimation of cycle time distribution with less computational effort. Lugaresi and Matta\cite{lugaresi2021automated} introduced a method for automatically extracting relevant features of a production system from event logs and aggregating activities that are not critical to system performance.
Our work comes closest to Lugaresi and Matta\cite{lugaresi2021automated} as we also use event logs as the only source of information for simulation model building. However, different from Lugaresi and Matta\cite{lugaresi2021automated}, we do not specify a priori objectives for guiding what level of detail to include in the simulation model. Instead, we use Inverse Reinforcement Learning (IRL) to learn how the system is controlled (i.e., dispatching policy), then use the learned value function to measure how similar machines are, and cluster similar ones together for state-space compression, guiding our resource aggregationin simulation model generation. To our knowledge, no previous research has integrated IRL into simulation-model generation process for digital twins.



\section{Problem Description}
\label{sec: problem}

A flexible flow shop (FFS) consists of $K$ sequential processing stages.
At each stage $k \in \{1,\ldots,K\}$, there is a set of unrelated parallel machines $M_k$ with stochastic processing times. Jobs arrive at stage 1, visit each stage sequentially, and leave the system after completing stage $K$. Due to limited capacity and time-varying workload, the system may experience congestion. Work-in-progress inventory is held in a buffer area that exists between each stage. A key operational component of the FFS system is the dispatching policy, which is responsible for assigning jobs to available machines $m$ and may differ at each stage\cite{joo2013dispatching}. We assume that the dispatching policy is not necessarily known, which is often the case in practice as operators may deviate from any pre-specified dispatching rules for certain reasons. The problem is how to represent the FFS system under arbitrary dispatching rules in a way that is learnable from event logs (see Table~\ref{tab:event_log_example}) and suitable for accurately simulating the system.


\section{Methodology}
\label{sec: method}
\subsection{MDP Formulation and State Abstraction}


We model dispatching in an FFS as a Markov decision process (MDP) inferred from event logs. A decision epoch occurs when a job is assigned to a machine, let $t$ denote the corresponding timestamp. From the event logs, we record the observable system state $X_t$ which includes current stage, queue length (local backlog relevant to the immediate dispatch decision) and work-in-process (total backlog at system level) at each decision epoch $t$; machine status is captured via the feasible action set, defined as the set of currently available machines in $M_{k}$. We define a candidate state abstraction level $\theta \in \Theta$, where $\Theta$ is a finite set of possible abstraction levels. For the illustration of our method in Section~\ref{sec: results}, we will restrict our attention  to a case with three distinct abstraction levels. In general, we specify the abstraction level of system state  $X_t$ by mapping it to an abstract state $s_t$ via a parameterized abstraction function $\psi_\theta$ such that $s_t = \psi_{\theta}(X_t)$.  The action $a_t \in M_{k}$ represents the machine selected for dispatching at decision epoch. The feasible set of machines $M_{k}$ depends on the current stage $k$. We estimate transition probabilities empirically from the event logs by counting transitions between consecutive decision epochs and normalizing the counts. We focus on the sequence of discrete decision epochs, forming state–action trajectories that reflect the dispatching behavior of FFS. That is, we reconstruct the dispatching process  as a Markov Decision Process (MDP) from the event logs. In Section~\ref{sec: IRL}, IRL will be used to learn the reward structure of the MDP that would lead to the observed event logs.

Selecting an appropriate level of state abstraction enables more stable learning in IRL while preserving the necessary system dynamics. For each candidate state abstraction level $\theta$, we used soft value iteration to calculate the value function $Q(s,a)$ under a provisional reward function. This yields a stochastic Boltzmann policy for the corresponding abstraction level. We then employed the learned policy to simulate the FFS and used negative log-likelihood (NLL) to measure the correctness of the actions simulated by this policy. Simulated key performance indicators such as throughput and system time are compared against those observed in the original event logs. Based on these two criteria, we selected the state abstraction level $\theta^\star$ with lowest KPI error and used it for subsequent IRL.

\subsection{IRL for Value-Based Resource Aggregation}
\label{sec: IRL}

Once the abstraction level $\theta^\star$ is determined in the previous section, we use the corresponding abstract-state trajectories $\{(s_t,a_t)\}$ to infer the dispatching policy of FFS. We apply the maximum entropy IRL framework and assume a linear reward function based solely on the state: $R(s) = w^\top \phi(s)$, where $w$ is the parameter vector and $\phi(s)$ is a feature vector derived from the abstract state.

Given a candidate parameter vector $w$, we use soft value iteration to compute the corresponding state–action value function $Q_w(s, a)$ and state value function $V_w(s)$. With Boltzmann policy, we compute candidate dispatching policy $\pi_w(a \mid s)$. $w$ is then estimated by iteratively updating to maximize the likelihood of the expert’s observed dispatching actions. With the learned value functions $Q(s, a)$ and $V(s)$ corresponding to $w$, we define the advantage function as $A(s,a) = Q(s,a) - V(s)$ which measures the relative benefit of selecting machine $a$ compared to the baseline value $V(s)$ in that state. 

We define the advantage vector $g(m)$ which is composed of $A(s,m)$ across different states (same abstraction level $\theta^\star$). Machines with similar advantage vector are considered to be behaviorally interchangeable and are therefore grouped into the same resource cluster by a clustering method ($k$-means).

\subsection{Evaluating Aggregation Schemes via Simulation}
Finally, we selected the optimal resource aggregation scheme through simulation-based evaluation. For each candidate clustering number, we constructed an aggregated simulation model where all machines in a cluster are represented as a single aggregated resource. The pooled resource capacity is defined as the total capacity of the machines in the cluster. The production process is then simulated using the learned policy. The simulation outputs (throughput and system time) are compared with the key performance indicators (KPIs) computed from the original event logs in the validation dataset. We select the smallest number of resource groups that can reproduce the observed system performance within an acceptable error range.

\section{Result}
\label{sec: results}
We evaluate the proposed method using synthetic FFS event logs generated by discrete-event simulation as illustrated in Table 1. The FFS consists of $K=3$ serial stages: 6 machines in stage 1, 8 machines in the stage 2 and 5 machines in the stage 3, with each machine having unit capacity. Inter-stage buffers are assumed to be infinite. Jobs arrive at stage 1 according to a Poisson process with exponentially distributed inter-arrival times of mean 8.0 time units. Each run releases $N=1200$ jobs and ends after all jobs have completed stage 3.

\begin{table}[!t]
\centering
\caption{Example event-log records from the synthetic flexible flow shop.}
\label{tab:event_log_example}
\scriptsize
\setlength{\tabcolsep}{4pt}
\begin{tabular}{r l r r l r r r r}
\toprule
time & event & job\_id & stage & machine\_id & stage\_buffer\_len & machine\_wait\_len & machine\_busy & WIP \\
\midrule
0.00 & queue\_enter & 0 & 1 & -- & 1 & 0 & 0 & 1 \\
0.00 & dispatch & 0 & 1 & \texttt{S1\_M1} & 1 & 0 & 0 & 1 \\
0.00 & service\_start & 0 & 1 & \texttt{S1\_M1} & 0 & 0 & 1 & 1 \\
2.59 & service\_end & 0 & 1 & \texttt{S1\_M1} & 0 & 0 & 1 & 1 \\
2.59 & queue\_enter & 0 & 2 & -- & 1 & 0 & 0 & 1 \\
2.60 & dispatch & 0 & 2 & \texttt{S2\_M1} & 1 & 0 & 0 & 1 \\
2.60 & service\_start & 0 & 2 & \texttt{S2\_M1} & 0 & 0 & 1 & 1 \\
25.63 & service\_end & 1 & 2 & \texttt{S2\_M2} & 0 & 0 & 4 & 7 \\
25.63 & queue\_enter & 1 & 3 & -- & 1 & 0 & 0 & 7 \\
25.70 & dispatch & 1 & 3 & \texttt{S3\_M1} & 1 & 0 & 0 & 7 \\
25.70 & service\_start & 1 & 3 & \texttt{S3\_M1} & 0 & 0 & 1 & 7 \\
30.27 & service\_end & 1 & 3 & \texttt{S3\_M1} & 0 & 0 & 2 & 8 \\
\bottomrule
\end{tabular}
\end{table}

Processing times are stochastic and stage dependent. At stage k, the basic processing time is sampled from a Gamma distribution parameterized by mean $\mu_k$ and coefficient of variation $\mathrm{CV}_k$, with $(\mu_1, cv_1) = (18, 0.6), (\mu_2, cv_2) = (22, 0.7)$ and $(\mu_3, cv_3) = (16, 0.5)$. Machine-specific processing time within the same stage is generated by multiplicative speed factors to the basic processing time. The speed factors are [0.8, 0.9, 1.0, 1.0, 1.2, 1.3] for stage 1, [0.8, 0.8, 0.9, 1.0, 1.0, 1.0, 1.3, 1.3] for stage 2 and [0.9, 1.0, 1.0, 1.2, 1.3] for stage 3, where smaller values indicate faster machines.

Stage-level dispatching is jointly determined by job selection rules and machine assignment rules. Stage 1 uses FIFO for job selection and a fastest available assignment policy which selects the available machine with the highest speed. Stages 2 and 3 use a Shortest Processing Time (SPT) job selection rule based on estimated stochastic processing times. Stage 2 assigns jobs to the least-loaded machine, whereas stage 3 follows the same fastest available logic as stage 1.

\begin{figure}[!t]
\centering
\setlength{\tabcolsep}{2pt}

\begin{minipage}[t]{0.62\linewidth}
\vspace{0pt}\centering
\begin{subfigure}[t]{0.49\linewidth}
\centering
\includegraphics[width=\linewidth, trim=0 0 0 24, clip]{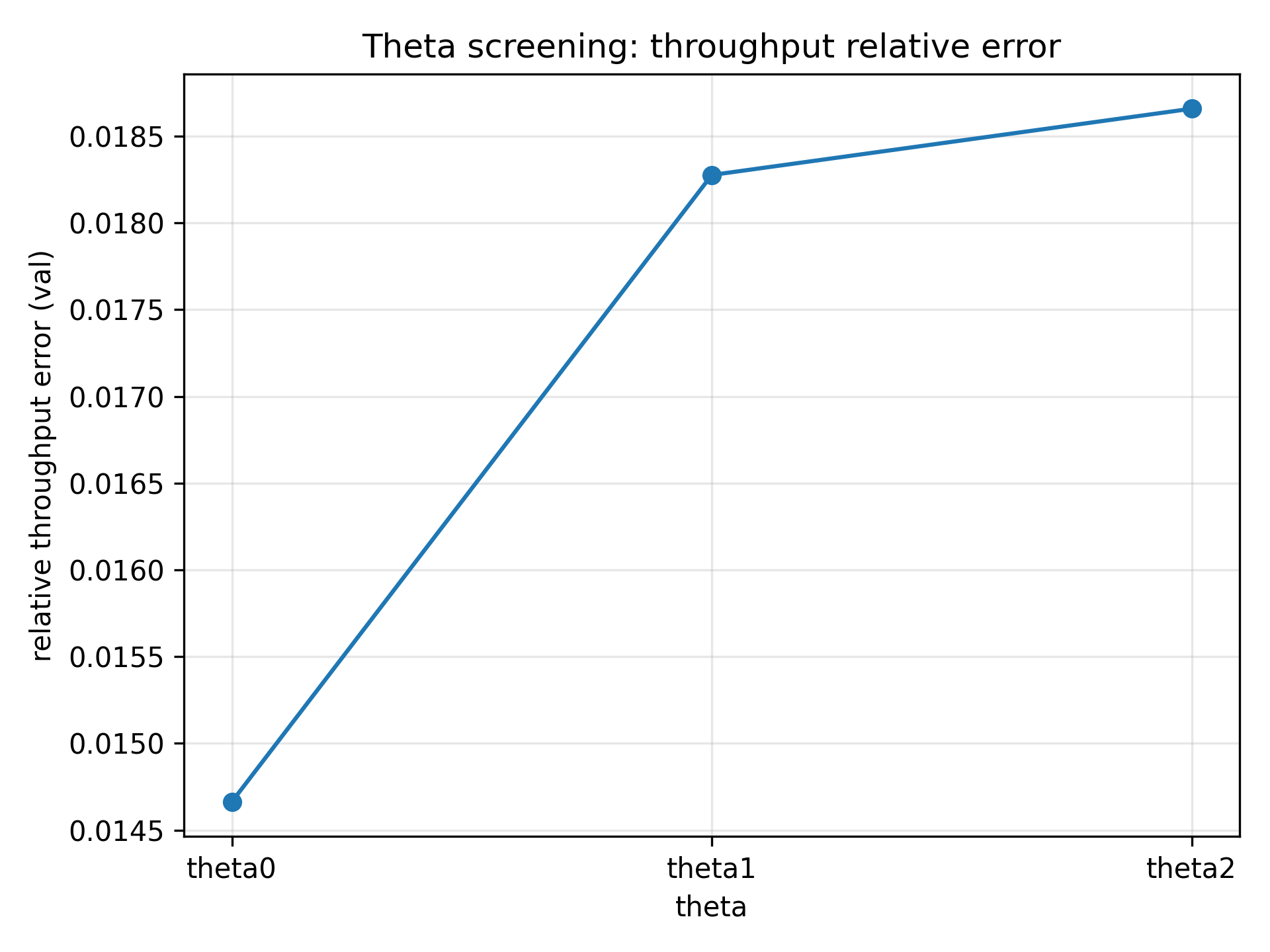}
\end{subfigure}\hfill
\begin{subfigure}[t]{0.49\linewidth}
\centering
\includegraphics[width=\linewidth, trim=0 0 0 24, clip]{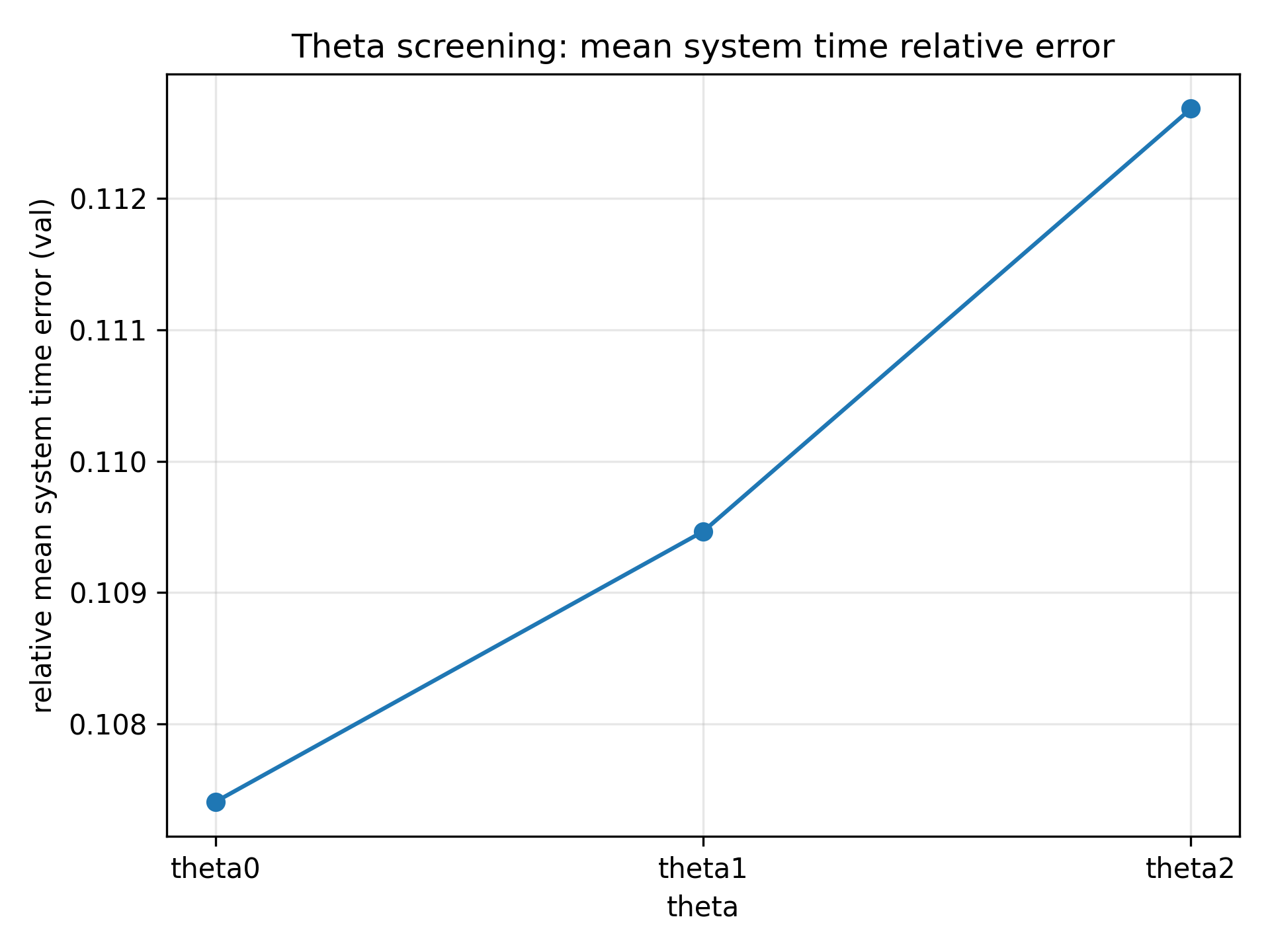}
\end{subfigure}
\vspace{-2pt}
\caption{Throughput and system-time relative errors.}
\label{fig:theta_errors}
\end{minipage}\hspace{3pt}
\begin{minipage}[t]{0.35\linewidth}
\vspace{0pt}\centering
\includegraphics[width=0.90\linewidth, trim=0 0 0 24, clip]{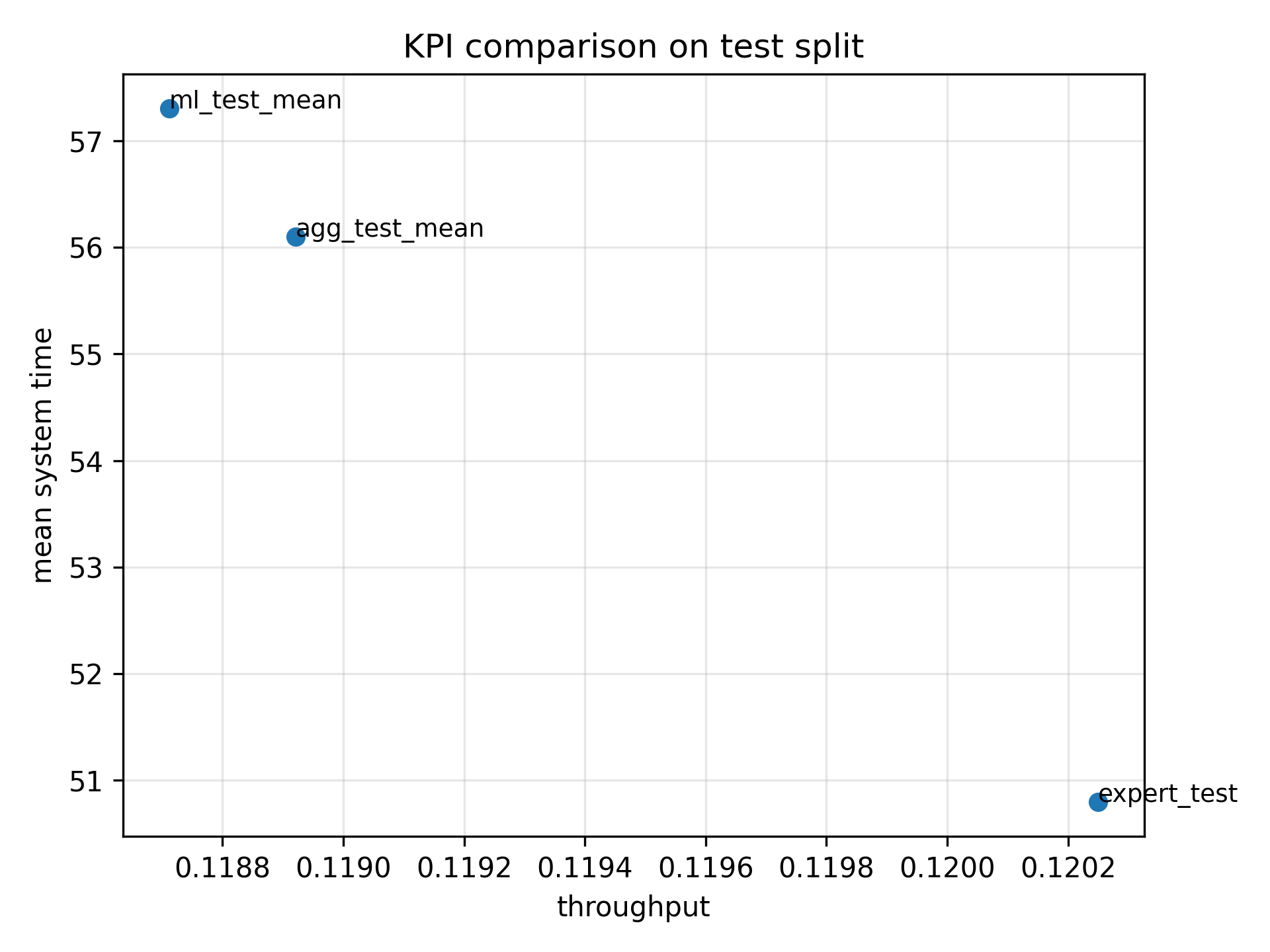}
\vspace{-2pt}
\caption{KPI comparison}
\label{fig:kpi_compare}
\end{minipage}

\end{figure}

The set of possible abstraction levels is defined as $\Theta = \{\theta_0, \theta_1, \theta_2\}$. The aggregate state in the first abstraction ($\theta_0$) includes the stage index and a binary indicator representing whether the queue is non-empty or not. The second abstraction ($\theta_1$) replaces the binary queue indicator with a discretized queue length interval, and the third abstraction ($\theta_2$) further adds a discretized work-in-process (WIP) interval. From Figure~\ref{fig:theta_errors}, on the validation set $\theta_0$ has the smallest relative error in both throughput and mean system time. $\theta_0$ also has a lower training NLL than $\theta_1$ and $\theta_2$ which means the model derived by $\theta_0$ is relatively more stable during learning. Based on these results, we set the state abstraction level $\theta^\ast = \theta_0$ in subsequent experiments to maximize scalability while preserving KPI accuracy.

\begin{figure}[!t]
\centering
\setlength{\tabcolsep}{2pt}

\begin{minipage}[t]{0.32\linewidth}
\vspace{0pt}\centering
\includegraphics[width=\linewidth, trim=0 0 0 24, clip]{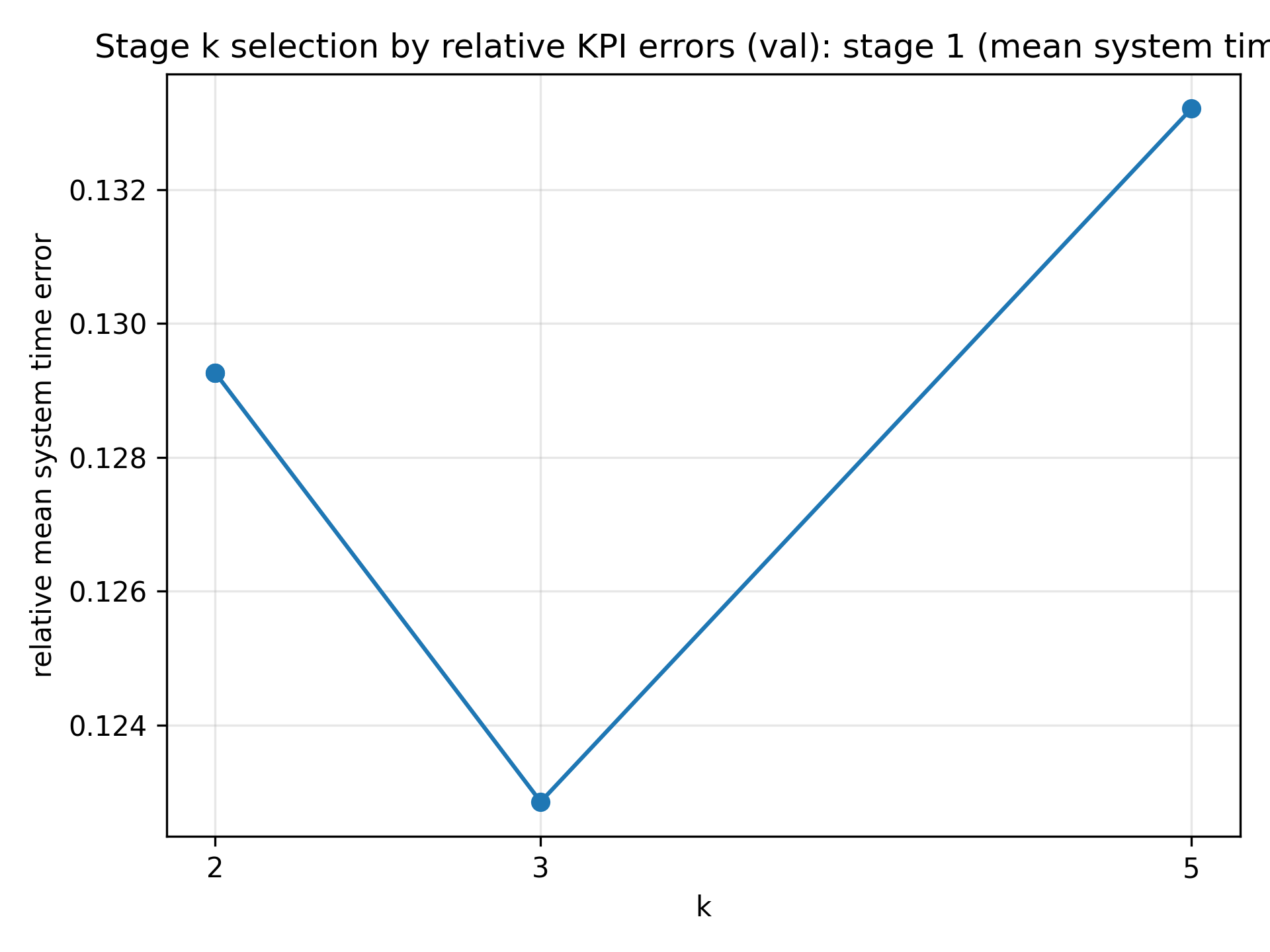}
\end{minipage}\hfill
\begin{minipage}[t]{0.32\linewidth}
\vspace{0pt}\centering
\includegraphics[width=\linewidth, trim=0 0 0 24, clip]{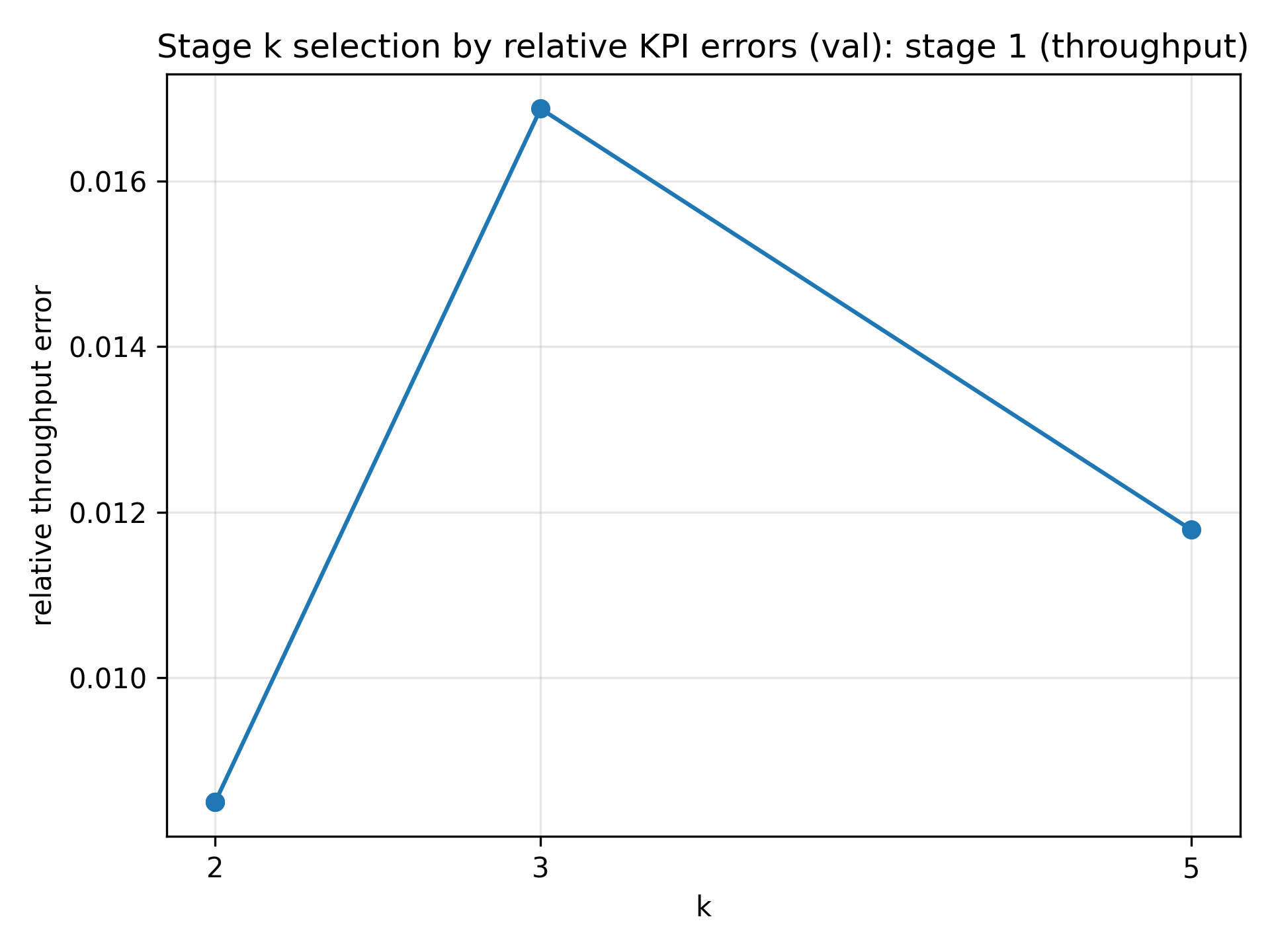}
\end{minipage}\hfill
\begin{minipage}[t]{0.32\linewidth}
\vspace{0pt}\centering
\includegraphics[width=\linewidth, trim=0 0 0 24, clip]{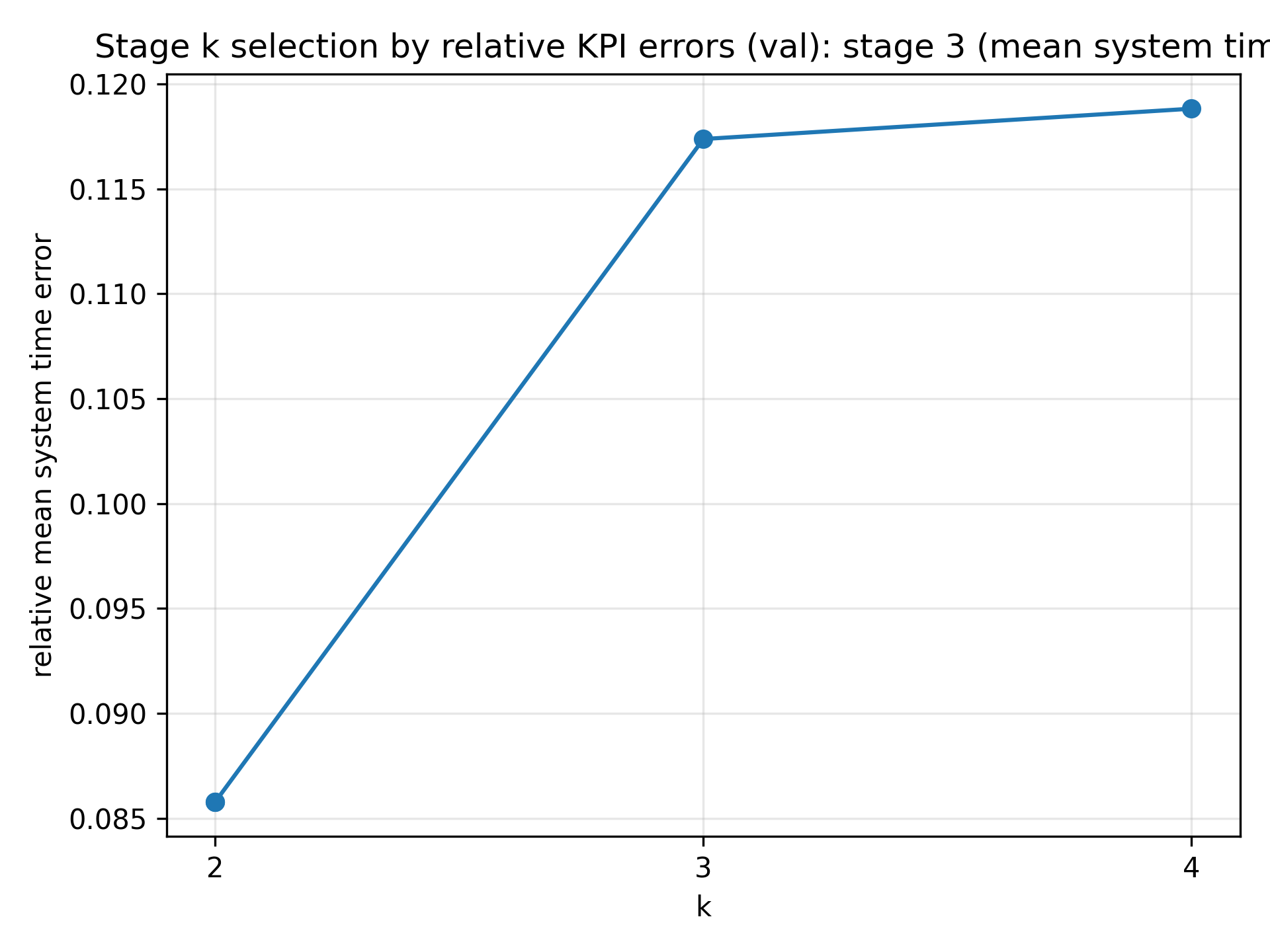}
\end{minipage}

\vspace{-4pt}
\caption{$k$ selection by ST for Stages 1--3.}
\label{fig:k_s}
\vspace{-6pt}
\end{figure}

\begin{figure}[!t]
\centering
\setlength{\tabcolsep}{2pt}

\begin{minipage}[t]{0.32\linewidth}
\vspace{0pt}\centering
\includegraphics[width=\linewidth, trim=0 0 0 24, clip]{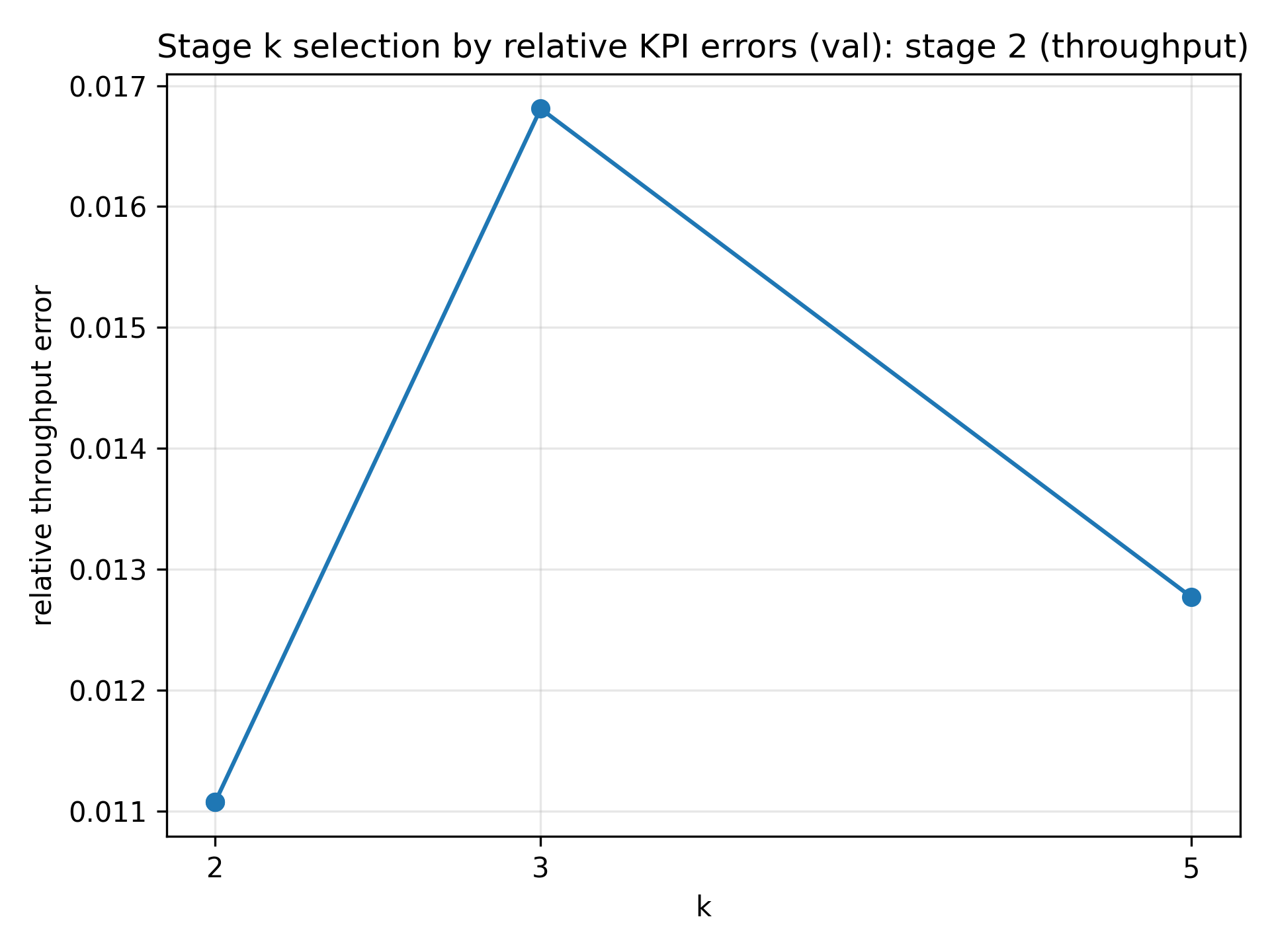}
\end{minipage}\hfill
\begin{minipage}[t]{0.32\linewidth}
\vspace{0pt}\centering
\includegraphics[width=\linewidth, trim=0 0 0 24, clip]{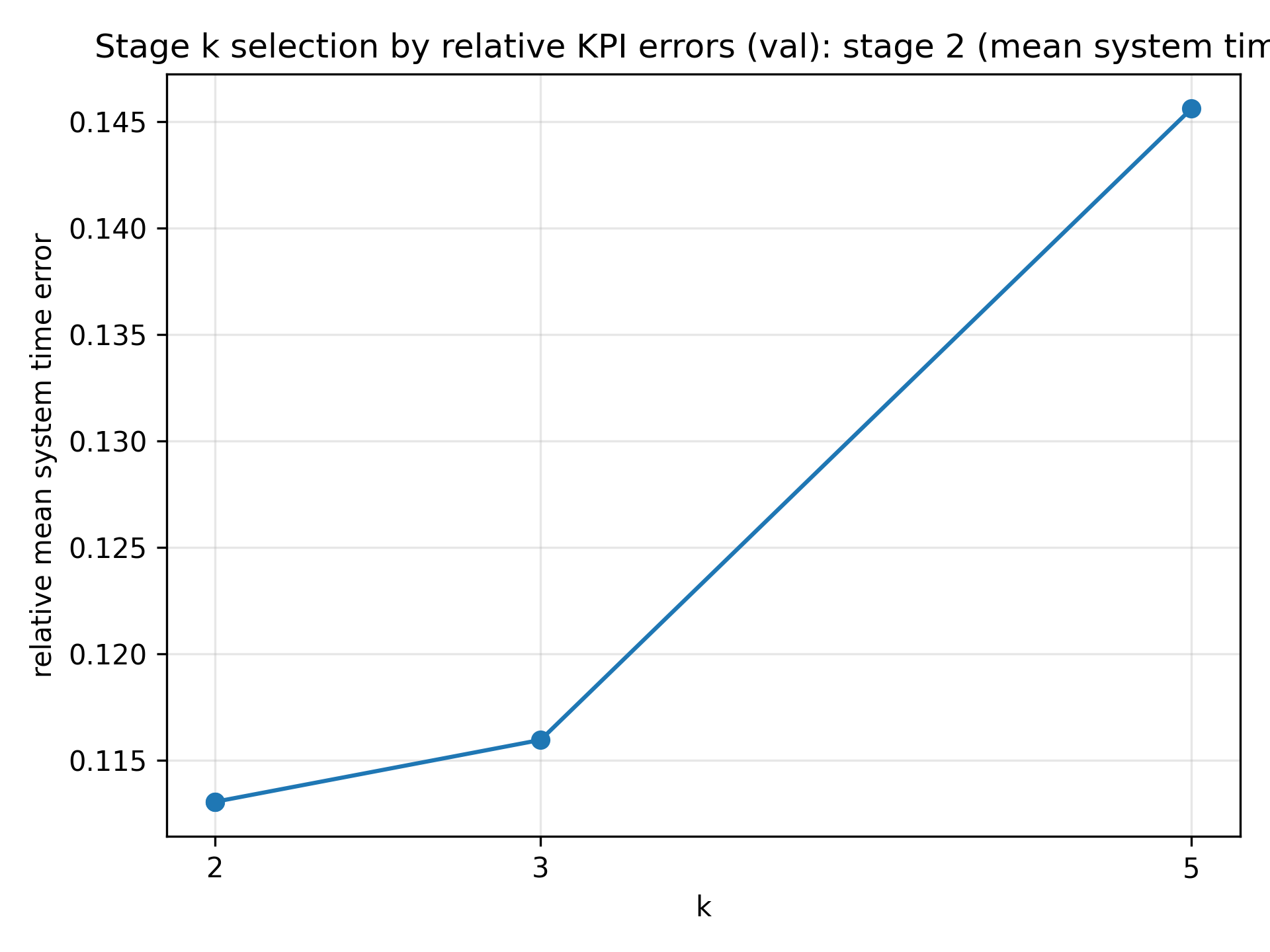}
\end{minipage}\hfill
\begin{minipage}[t]{0.32\linewidth}
\vspace{0pt}\centering
\includegraphics[width=\linewidth, trim=0 0 0 24, clip]{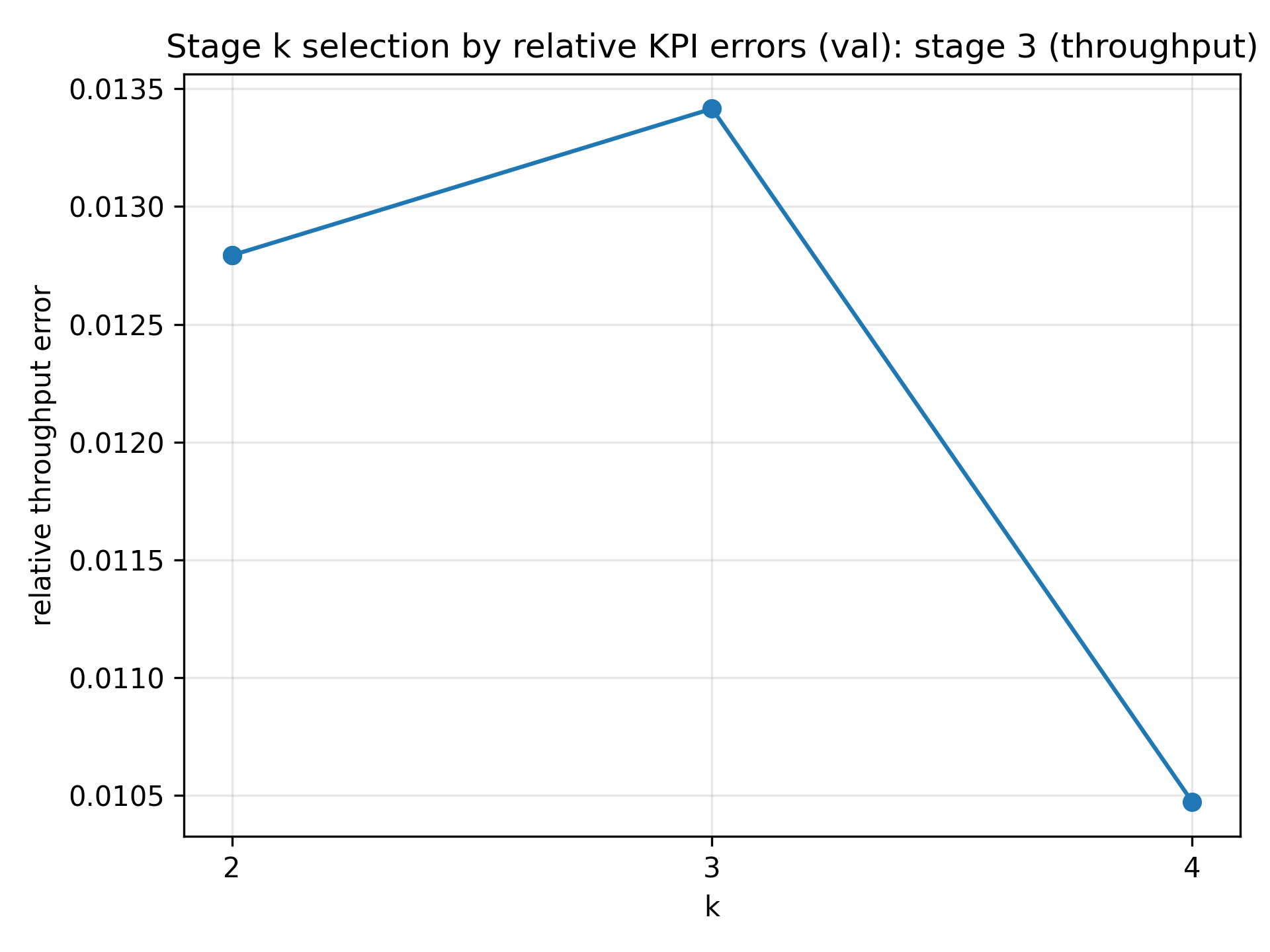}
\end{minipage}

\vspace{-4pt}
\caption{$k$ selection by throughput for Stages 1--3.}
\label{fig:k_thr}
\vspace{-6pt}
\end{figure}

Leveraging the machine advantage representation derived from IRL, we cluster machines using $k$-means and select $k$ by minimizing the validation KPI error of the aggregated simulation. As shown in Figures~\ref{fig:k_s} and~\ref{fig:k_thr} , for Stage 1 we choose $k = 2$, as it achieves the lowest validation throughput error while maintaining competitive mean system time error. Cluster 1 includes $m_4$ and $m_6$ and Cluster 2 includes $m_1$, $m_2$, $m_3$ and $m_5$. For Stages 2, 3, $k = 2$ achieves the best throughput and system time errors. In addition, we isolate rarely used machines (defined as those dispatched fewer than 20 times in the event log) into a separate cluster. Therefore, the final number of clusters for Stage 2 is 3. In stage 2, Cluster 1 includes $m_1$ and $m_3$, Cluster 2 includes $m_2$, $m_4$, $m_5$ and $m_6$ and Cluster 3 includes $m_7$ and $m_8$. In stage 3, Cluster 1 includes $m_1$, $m_2$ and $m_5$ and Cluster 2 includes $m_3$ and $m_4$.

We assess the fidelity of the aggregated model by comparing three sets of KPIs: expert KPIs computed from the expert event logs, machine-level simulation KPIs, and aggregated simulation KPIs under the learned clustering scheme. As shown in Table~\ref{tab:kpi_comparison}, the aggregated model closely matches the machine-level model, with relative errors of 0.18\% in throughput and 2.09\% in mean system time, indicating that the aggregated simulation faithfully reproduces the machine-level system behavior on these two key metrics. Figure~\ref{fig:kpi_compare} provides a clearer visualization of this result. Furthermore, using the expert KPIs as the reference, the machine-level and aggregated models achieve relative throughput errors of 1.28\% and 1.11\% respectively, and relative mean system time errors of 12.80\% and 10.43\% respectively. Therefore, relative to the machine-level model, the aggregated model shows a slight improvement in matching the event logs.

\begin{table}[!t]
\centering
\caption{KPI comparison across expert (event logs), machine-level simulation, and aggregated simulation.}
\label{tab:kpi_comparison}
\small
\setlength{\tabcolsep}{3pt}
\renewcommand{\arraystretch}{0.88}
\begin{tabular}{lcc}
\toprule
\textbf{Method} & \textbf{Throughput} & \textbf{Mean system time} \\
\midrule
Expert (event logs) & 0.12025 & 50.80 \\
Machine-level simulation & 0.11871 & 57.30 \\
Aggregated simulation & 0.11892 & 56.10 \\
\bottomrule
\end{tabular}
\end{table}

\section{Conclusion}
\label{sec: conc}
This paper proposes a resource aggregation method based on inverse reinforcement learning to compress machine-level dispatching behavior into a compact yet accurate simulation model, illustrated using a flexible flow shop manufacturing system. 
Building on the dispatching policy learned via inverse reinforcement learning, we introduce a value-based machine representation and a performance-driven clustering choosing approach to aggregate resources, thereby significantly reducing model complexity. The resulting aggregated simulation model preserves key system-level performance indicators while significantly decreasing model size and computational burden.

Despite these encouraging results, some limitations remain. First, the inverse reinforcement learning module adopts a linear and state-only reward function. While this design improves stability, it may be insufficient to capture more complex policies driven by differences in job types or machine-specific performance. Second, our experiments are limited to a flexible flow shop environment and primarily evaluate throughput and system time. Additional validation across broader manufacturing structures and a wider range of key performance indicators is needed to assess generalizability and clarify when value-based aggregation methods are most effective.

\bibliographystyle{IEEEtran}
\bibliography{references}
\end{document}